\documentclass[runningheads]{llncs}
\usepackage[T1]{fontenc}
\usepackage{siunitx}
\usepackage{amsmath}
\usepackage{amsfonts}
\usepackage{amssymb}
\usepackage{bm}
\usepackage{scalerel}

\usepackage{subcaption}
\usepackage{hyperref}
\usepackage{multirow}%
\usepackage{mathrsfs}%
\usepackage[title]{appendix}%
\usepackage{textcomp}%
\usepackage{manyfoot}%
\usepackage{booktabs}%
\usepackage{algorithm}%
\usepackage{algorithmicx}%
\usepackage{algpseudocode}%
\usepackage{listings}%

\usepackage{lineno}

\def\app#1#2{%
  \mathrel{%
    \setbox0=\hbox{$#1\sim$}%
    \setbox2=\hbox{%
      \rlap{\hbox{$#1\propto$}}%
      \lower1.1\ht0\box0%
    }%
    \raise0.25\ht2\box2%
  }%
}

\begin{document}

\title{Emotion in an active inference model of human driving}

\author{Julian F. Schumann\inst{1} \and
Johan Engström\inst{2} \and
Ran Wei\inst{2} \and
Jens Kober\inst{1} \and
Martijn Wisse\inst{1} \and
Arkady Zgonnikov\inst{1}}
\authorrunning{Schumann et al.}
%
\institute{Department of Cognitive Robotics, Delft University of Technology\\
\email{\{j.f.schumann, j.kober, m.wisse, a.zgonnikov\}@tudelft.nl} \and
Waymo LLC, Mountain View, CA, USA\\
\email{\{jengstrom,ranweii\}@waymo.com}}


\maketitle
\begin{abstract}
Active inference has emerged as a principled framework for modeling adaptive behavior by balancing goal-directed action with uncertainty reduction. It has been successfully applied across biological and artificial systems, including recent work on human driving. However, existing active inference models of driving have yet to address an important determinant of behavior in traffic: affective state, which significantly influences decision-making. Prior work in non-traffic domains has explored active inference agents in which emotions are represented along the axes of valence and arousal in the circumplex model. However, this work has been limited to simplified settings with discrete state spaces.
In this work, we propose an expanded formulation of valence and arousal that can be extracted from a more complex active inference model of driving with continuous states. In particular, we condition affective estimates not only on the current state but also on predicted future outcomes. We evaluate the proposed approach in two interactive driving scenarios and show that the resulting emotion signals correspond to affective patterns reported in similar scenarios.

\end{abstract}

\section{Introduction}
Active inference is a first-principles account of adaptive behavior of organisms (usually referred as agents) selecting actions which minimize their expected free energy and thereby their surprise~\cite{friston_active_2017,parr_active_2022}. This balances the drive to achieve preferred outcomes (maximize pragmatic value) with the drive to reduce uncertainty about the environment (maximize epistemic value).
Active inference has been applied across various levels of biological organization, spanning from modeling single cells \cite{friston_life_2013} to adaptive processes in plants~\cite{calvo_predicting_2017} and models of human cognition and behavior~\cite{van_de_maele_hierarchical_2024}.

Recently, active inference has also been applied to the modeling of human driving, including scenarios such as car-following~\cite{wei_active_2023}, lane-keeping~\cite{engstrom_resolving_2024}, collision avoidance~\cite{schumann_active_2025}, and interactions at intersections~\cite{schumann2026resolving}. Here, active inference offers a general yet robust and mechanistically interpretable framework~\cite{schumann_active_2025} that provides a potential middle ground between traditional mechanistic behavioral models~\cite{svard_computational_2021,zgonnikov_should_2022,schumann_using_2023,siebinga_model_2024-1} and ``black-box'' data driven, machine-learning-based, approaches to road user behavior modeling~\cite{rowe_fjmp_2023,meszaros_trajflow_2024,nayakanti_wayformer_2023}. In addition to providing a deeper fundamental understanding of human behavior in high-stakes dynamic settings, such models also have practical applications. For instance, they can be used in simulation-based validation and testing pipelines for autonomous vehicles (AV)~\cite{fremont2020formal,montali_waymo_2023} or as human baselines to compare AVs against~\cite{olleja2025validation}.

An important aspect of human driving behavior -- so far unaccounted for in these active inference models -- is the role of emotions. Previous work suggests that human behavior and driving capability is highly influenced by affective states~\cite{vaa2007modelling,summala2007towards,pecher2011influence,steinhauser2018effects}. Consequently, accounting for the emotional state of human drivers and its underlying causes in real world-scenarios is a key area for further development aspect of existing active inference models of human driving. While several existing active inference-based models of emotion have been proposed in on-traffic domains models~\cite{joffily2013emotional,smith2019simulating,hesp2021deeply,pattisapu2024free} these are limited to simplified toy problems with discrete state spaces and lack the capacity for realistic planning, which is generally needed to model human road user behavior in  most traffic scenarios.

In this paper, we extend the previous work of Pattisapu et al. on the modeling of emotional states~\cite{pattisapu2024free} in a simplified discrete task (searching for a lost item) to an existing active inference model of driving~\cite{engstrom_resolving_2024,schumann_active_2025,schumann2026resolving}. Specifically, we generalize the formulation to a continuous state space and account for policies of actions planned over multiple time steps. 
We then evaluate the proposed approach in two experimental scenarios. Our results indicate that the valence and arousal signals align with results reported previously in similar scenarios, suggesting that active inference provides a viable and principled framework for modeling affective components of human driving behavior.

\section{Background}
\subsection{Emotions}\label{sec:Emotion_definition}
While some theories treat emotions such as anger and happiness as discrete categories~\cite{ekman2011meant,lindquist2012brain}, others adopt a continuous representation. A prominent example is Russell’s circumplex model of affect~\cite{russell1980circumplex}, which places emotions along the dimensions of valence and arousal. This approach avoids debates about the exact number of emotions~\cite{ortony2022all} and accommodates overlapping affective states~\cite{dutton1974some}. It has also been applied to driving contexts~\cite{samuel2019riding,du2020examining}, motivating its use here.

Within active inference, internal beliefs and their uncertainties provide a natural substrate for modeling affect~\cite{joffily2013emotional,smith2019simulating}. Early work mainly focused on valence, relating it to the derivative of free energy~\cite{joffily2013emotional} or changes in confidence over predictions and actions~\cite{hesp2021deeply}. Building on this, Pattisapu et al.~\cite{pattisapu2024free} mapped active inference quantities onto both dimensions of the circumplex model. In their formulation, \emph{valence} -- the degree of positive or negative affect -- corresponds to the difference between expected and realized utility, while \emph{arousal} --  reflecting the level of alertness or activation -- is given by the entropy of the agent’s posterior belief. Positive valence thus reflects a pessimistic prediction error, consistent with biological links to dopamine release~\cite{schultz2016dopamine}, whereas arousal is interpreted as related to uncertainty, consistent with prior work relating it to amygdala activity~\cite{wilson2013neural,li2014conditions,barrett2017theory,corcoran2019allostatic}.

\subsection{An Active Inference Model of Human Driving}\label{sec:Driving_model}
In this paper, we start from the active inference model of human driver collision avoidance behavior presented in Schumann et al.~\cite{schumann_active_2025} and previous work~\cite{engstrom_resolving_2024,schumann2026resolving} and extend it with a model of driver emotion.
As in the previous models, we assume an external \emph{generative process} that generates observations $\bm{o}$, which is approximated by the agent’s internal \emph{generative model} operating on the stochastic belief $q(\bm{s})$ (represented as particle filters), with the observation probability $p(\bm{o}\vert\bm{s})$ and the transition likelihood $p(\bm{s}'\vert\bm{s},\bm{a})$. This transition model has two components: (1) A kinematic likelihood based on the bicycle model~\cite{polack2017kinematic}, which biases the model to assume other agents to maintain a constant velocity and steering angle and (2) a normative likelihood, which biases the model to assume that other agents comply with common traffic norms (such as lane following or priority rules). 
An agent iteratively selects actions (acceleration and steering rate) based on a multi-step process, described in further detail in~\cite{schumann_active_2025}:
\begin{enumerate}
    \item At time $t$, the agent makes an observation $\bm{o}_t \sim \widehat{p}(\bm{o}_t\vert\bm{\eta}_t)$ and uses it to find its updated belief $q(\bm{s}_t)$ using Bayesian inference:
    \begin{equation}
        q(\bm{s}_{t}) \propto p(\bm{o}_{t}\vert \bm{s}_{t}) \, \mathbb{E}_{q(\bm{s}_{t-1})} p(\bm{s}_{t}\vert \bm{s}_{t-1}, \bm{a}_{t-1})\,.
    \end{equation}
    \item The agent generates candidate policies $\bm{\pi}_t = \{\bm{a}_{\tau} \vert \tau \in \{t,\hdots,t+H-1\}\}$. These are sequences of actions $\bm{a}$ with prediction horizon $H$.
    \item For each candidate policy, the \emph{generative model}'s state transition function is used to form a belief about the future states corresponding to that respective policy. Specifically, starting with $\widetilde{q}(\bm{s}_{t}) = q(\bm{s}_t)$, we generate 
    \begin{equation}
        \widetilde{Q}(\bm{\pi}_t) = \{\widetilde{q}(\bm{s}_{\tau}) = \mathbb{E}_{\widetilde{q}(\bm{s}_{\tau-1})} p(\bm{s}_{\tau}\vert\bm{s}_{\tau-1},\bm{a}_{\tau-1}) \mid \tau \in \{t+1,\hdots,t+H\}\} \,.
    \end{equation}
    \item Candidate policies are evaluated based on the expected free energy $G$ (EFE), taking into account behavioral predictions of other road user's behavior,and the policy with the lowest EFE is selected. The EFE scores both the value of achieving goals (pragmatic value) and the value of obtaining new information to resolve uncertainty (epistemic value), and can be defined as
    \begin{equation}
        G(\bm{\pi}_t) = - \sum\limits_{\widetilde{q}(\bm{s}) \in \widetilde{Q}(\bm{\pi}_T)} \underbrace{\mathbb{E}_{\widetilde{q}(\bm{s})} \mathbb{E}_{p(\bm{o}\vert\bm{s})}p(\bm{o})}_{g_{\text{prag}}(\widetilde{q}(\bm{s}))} + \underbrace{\mathcal{H}\left(\mathbb{E}_{\widetilde{q}(\bm{s})} p(\bm{o}\vert \bm{s})\right) - \mathbb{E}_{\widetilde{q}(\bm{s})} \mathcal{H}\left( p(\bm{o}\vert \bm{s})\right)}_{g_{\text{epist}}(\widetilde{q}(\bm{s}))}\,.
    \end{equation}
    Here, $p(\bm{o})$ is the preference prior encoding an agents desired observations.
    \item To model realistic reaction times, the agent only updates its currently pursued policy if the accumulated surprise $E_t = E_{t-1} + \lambda \varepsilon_t$ against the current policy has reached a predefined threshold, with
    \begin{equation}
        \varepsilon_t =  H \underset{\widetilde{q}(\bm{s})}{\max} \, g_{\text{prag}}(\widetilde{q}(\bm{s})) - \sum\limits_{\widetilde{q}(\bm{s}) \in \widetilde{Q}(\bm{\pi}_t)} g_{\text{prag}}(\widetilde{q}(\bm{s})) \, .
    \end{equation}
    Otherwise, it extends the current reference policy $\bm{\pi}^R_t$ by appending a single action at the end of the time horizon. The result of this process is the chosen policy $\bm{\pi}^{*}_t$.
\end{enumerate}
The first action $\bm{a}_t$ of $\bm{\pi}^{*}_t$ is then passed to and updates the environment (\emph{generative process}), and the cycle is repeated.

\section{Modeling Emotions in Human Driving}
As discussed in Section~\ref{sec:Emotion_definition}, Pattisapu et al.~\cite{pattisapu2024free} defined emotions based on a simple model with discrete states. In mathematical terms, valence $V$ and arousal $A$ are defined as
\begin{equation}\label{eq:Valence_old}
    V_t = \underbrace{g_{\text{prag}}(q(\bm{s}_t))}_{\text{Actual Value }V^{(A)}} - \underbrace{g_{\text{prag}}(\widetilde{q}(\bm{s}_t))}_{\text{Expected Value }V^{(E)}} \, \text{and} \, \, A_t = \mathcal{H}(q(\bm{s}_t)) \,,
\end{equation}
where $\widetilde{q}(\bm{s}_t)$ is the first element in $\widetilde{Q}(\bm{\pi}^{*}_{t-1})$.
In the following sections, we extend these definitions to the continuous state space model for human driving described in Section~\ref{sec:Driving_model}

\subsection{Valence} 
In the original emotion model~\cite{pattisapu2024free} from Equation~\eqref{eq:Valence_old}, the different recorded time steps $t$ do not have any inherent frequency. However, in our scenario, the state transition function always moves forward the state by the constant time $\Delta t$. Consequently, to make sure the emotions are not dependent on the time step, we propose to divide the Valence calculation by this factor.
Additionally, in our model, we use a prediction horizon of $H > 1$, our proposed implementation -- here defined as $\mathcal{V}$ -- does take this into account. This corresponds to the appraisal theory of emotion~\cite{scherer2001appraisal}, which posits that emotions can be influenced by a person's assessments of the likelihood of future events. Additionally, given the random nature of the particle filter in the underlying driver behavior model, we also aimed to formulate a more stable version of valence. Consequently, the expected value $\mathcal{V}^{(E)}$ considers the pragmatic value over the predicted states associated with the previous $N=5$ policies, with 
\begin{equation}\begin{aligned}
    \mathcal{V}_t^{(E)} = \frac{1}{N H_N\Delta t} \sum_{n=1}^N \sum\limits_{\widetilde{q}(\bm{s}) \in \widetilde{Q}_n\left(\bm{\pi}^{*}_{t-n}\right)} g_{\text{prag}}(\widetilde{q}(\bm{s}))  \,.
\end{aligned}
\end{equation}
Here, we define $\widetilde{Q}_n(\bm{\pi}^{*}_{t-n}) = \{\widetilde{q}(\bm{s}_\tau) \in \widetilde{Q}(\bm{\pi}^{*}_{t-n}) \mid \tau \in \mathcal{T}\}$, where $\mathcal{T} = \{t,\ldots,t+H-N\}$ is the common prediction horizon shared by all considered $\widetilde{Q}(\bm{\pi})$. Hence, all $\widetilde{q}(\bm{s}_\tau)$ with $\tau \notin \mathcal{T}$ are removed (e.g., $\widetilde{q}(\bm{s}_{t+H-1}) \notin \widetilde{Q}_1(\bm{\pi}_{t-1})$ since, for instance, $\widetilde{Q}(\bm{\pi}_{t-N})$ does not extend that far into the future). 
Using a similar approach for the actual value $\mathcal{V}^{(A)}$, we get
\begin{equation}
    \mathcal{V}_t^{(A)} =  \frac{1}{H_N\Delta t} \left( g_{\text{prag}}(q(\bm{s}_t)) + \sum\limits_{\widetilde{q}(\bm{s}) \in Q\left(\bm{\pi}^{R}_{t}\right)} g_{\text{prag}}(\widetilde{q}(\bm{s})) \right)\,,
\end{equation}
with $Q\left(\bm{\pi}^{R}_{t}\right) = \left\{\widetilde{q}(\bm{s}_{\tau})  \in \widetilde{Q}\left(\bm{\pi}^{R}_{t}\right) \mid \tau \in \mathcal{T}\setminus \{t\} \right\}$.
Importantly, we calculate the actual value based on the reference policy $\bm{\pi}^R_t$ (i.e., the policy currently still in effect). This is consistent with ideas of valence as a reflection of ongoing organism-environment interactions~\cite{frijda1986emotions,russell2003core}, preventing unimplemented counterfactual policies from inducing affective changes.
Combining the actual and expected value then results in the new valence term
\begin{equation}
    \mathcal{V}_t = \mathcal{V}_t^{(A)} - \mathcal{V}_t^{(E)}
\end{equation}

\subsection{Arousal}
We also have to adapt the calculation of arousal. In principle, one could expand the original definition of arousal in Equation~\eqref{eq:Valence_old} by including the uncertainty about not only the current kinematic state, but also the future state, assuming that the modeled agent generally reasons over this longer prediction horizon. This inclusion of uncertainty about future events is again consistent with appraisal theory~\cite{scherer2001appraisal}. 
However, this approach breaks down upon closer examination, as linking kinematic uncertainty directly to arousal can result in irrational results. For instance, in models including looming-based perception~\cite{schumann_active_2025}, a nearby agent would be perceived more precisely, thereby reducing the associated kinematic uncertainty and, consequently, the resulting arousal. This outcome is highly counterintuitive, as it implies that arousal would be highest for distant agents that have no realistic chance of interacting with the modeled agent, whereas a vehicle on an impending collision course would evoke lower arousal\footnote{A more detailed discussion, along with simulation results using kinematic arousal, can be found at \url{https://osf.io/7x53c/files/m29qe}.}.

Consequently, rather than naively generalizing the original mathematical construction, we instead build on the underlying semantic interpretation proposed by Pattisapu et al.~\cite{pattisapu2024free}, computing arousal over meaningful abstract states (e.g., the presence of a key at a particular location). In particular, we can extract such abstract states from our predicted belief states. While the number of potential behavior categories in traffic problems is unbounded, we decided to focus on the predicted likelihood of collision $p_{\text{coll}}$. This is not only supported by the strong relation between endangerment and arousal shown in the literature~\cite{cannon1929bodily} (i.e., this uncertainty is actually relevant to the agent), but is also generally applicable, as it is not dependent on certain scenarios or environments. Mathematically, we define that
\begin{equation}
    p_{\text{coll},t} = \mathbb{E}_{\widetilde{Q}(\bm{\pi}^{*})} \underset{\tau \in \{t+1, \ldots, t+H\}}{\max} \bm{1}_{\text{coll}}\left(\bm{s}_{\tau} \right) \,, 
\end{equation}
where $\bm{1}_{\text{coll}}$ is a binary collision classifier based on the separating axis theorem~\cite{liang2015research}.
Given the resulting behavioral state of $\bm{s}_B=\{p_{\text{coll}}, 1- p_{\text{coll}}\}$, Equation~\eqref{eq:Valence_old} then resolves into the binary entropy function
\begin{equation}
    \mathcal{A}_{t} = -\frac{1}{2}\left(p_{\text{coll},t} \ln(p_{\text{coll},t}) + (1-p_{\text{coll},t}) \ln(1-p_{\text{coll},t})  \vphantom{1^1}\right) \in [0, \ln(2)]\,.
\end{equation}

\section{Experiments}
After describing our method for modeling emotions based on the internal states of an active inference model, we apply this to two scenarios. In each scenario, we run 128 simulations, where in one half the modeled agent interacts with a compliant driver, while the other half studies instances of adversarial behavior of another agent. We will both study the aggregate emotions in the simulations as well as some examples described in more detail.
\subsection{Scenarios}

\noindent\textbf{Lateral incursion.}
This scenario includes a road with two opposing lanes, with the modeled agent $A$ driving rightwards along the road at 40 miles per hour (i.e., $\SI{17.88}{m.s^{-1}}$). The oncoming agent $B$, initially at a distance of $\SI{200}{m}$ in the opposite lane approaches agent $A$ with the same velocity. Once it comes close enough, it executes one of two predefined policies: (1) \textit{Compliant}: It drives with constant speed along the center of its lane until passing agent $A$ or (2) \textit{Non-compliant}: It executes a lateral incursion maneuver that puts it on a collision course with agent $A$. This scenario adapted from a driving simulator study by Johnson \emph{et al.}~\cite{johnson2025looking}; here we used the \emph{steep} incursion version of the maneuver, as described in~\cite{schumann_active_2025}.

\noindent\textbf{Intersection with priority violation.}
This scenario focuses on an orthogonal intersection of two one-lane roads, previously studied in Schumann et al.~\cite{schumann2026resolving}. The modeled agent $A$ approaches the intersection along the priority road at an initial distance of $\SI{65}{m}$ and a velocity of $\SI{10}{m.s^{-1}}$. It has a lead of $\SI{3.25}{m}$ over agent $B$, which is approaching the intersection at the same speed. We again consider two different variants: (1) \textit{Compliant}: When approaching the intersection, agent $B$ -- consistent with the ``give way'' sign -- starts to brake after $\SI{2}{s}$. (2) \textit{Non-compliant}: Agent $B$ maintains its current speed until reaching the intersection, ignoring the traffic rules.

\subsection{Results}
\subsubsection{Lateral incursion.}
\begin{figure*}[!ht]
    \centering
    \includegraphics[scale=0.9]{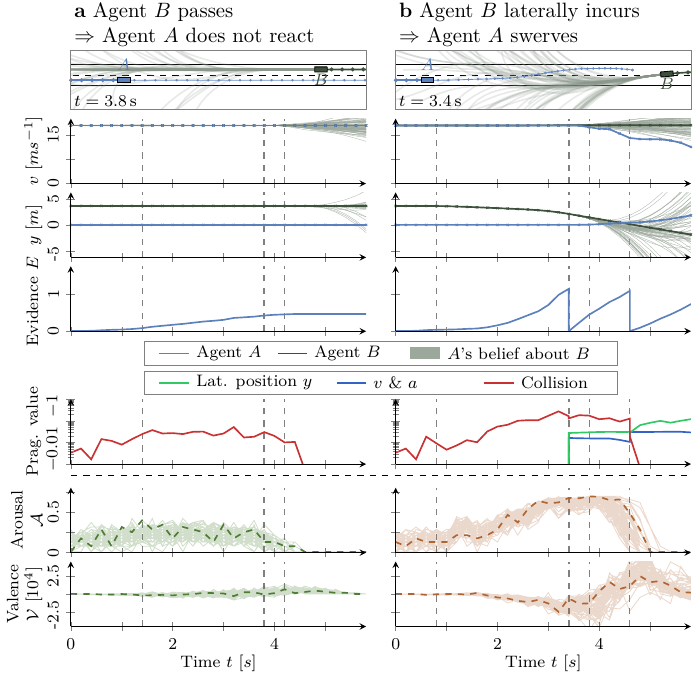}
    \caption{Examples of two possible outcomes in the lateral incursion scenario. The thin lines in the plot of $v$ and $y$ correspond to the prediction shown in the top plots. \textbf{a}) The compliant agent $B$ stays in its lane, allowing agent $A$ to pass without the need for any adjustment. \textbf{b}) The non-compliant agent $B$ incurs into the lane of agent $A$, which is forced to avoid a collision by moving into the opposite lane itself. In the emotion plots (lowest two panels), the dashed lines correspond to the examples shown in the panels above, while the thin lines represent the other simulated case. The dashed vertical lines correspond to key points in time discussed in the text.}
    \label{fig:oncoming_examples}
    \vspace{-0.5cm}
\end{figure*}
In the lateral incursion scenario, the emotions of agent $A$ differ significantly depending on the behavior of agent $B$. In the scenario where agent $B$ is compliant (Figure~\ref{fig:oncoming_examples}\textbf{a}), we see an initial increase in arousal $\mathcal{A}$ as the two agents approach each other. This is driven by the gradually increasing likelihood of a collision (see the collision component in the pragmatic value)\footnote{The normative bias against the oncoming agent leaving its lane makes this event less likely, but it is not considered impossible. Additionally, given the nonlinear nature of the binary entropy function underlying $\mathcal{A}$, even minuscule collision likelihoods can result in noticeable arousal.}, peaking at around $t=\SI{1.4}{s}$ (first dashed line). Afterwards, as the agents get closer, the predicted likelihood of collisions -- and correspondingly $\mathcal{A}$ -- decreases again. Particularly, at this shorter distance, predicted incursions by agent $B$ occur too late to still affect agent $A$, as can be seen in the top panel of Figure~\ref{fig:oncoming_examples}\textbf{a}. This reduction in collision likelihood, with agent $B$ behaving more benignly than expected, leads to an increasing pragmatic value and consequently to a positive valence, which peaks at around $t=\SI{4.2}{s}$ (third dashed line). 

There are marked differences in the adversarial setting (Figure~\ref{fig:oncoming_examples}\textbf{b}). Specifically, after $t=\SI{0.8}{s}$ (first dashed line), when agent $B$ begins its incursion, we observe a consistent increase in $\mathcal{A}$. This is a consequence of the norm-violating nature of the maneuver: agent $A$ loses trust in the norm compliance of agent $B$ and no longer applies norm conditioning in its predictions, leading to higher uncertainty. As a result, agent $A$ takes the possibility of a collision more seriously over the following seconds than in the scenario with the compliant other agent, as reflected in the faster increase of both the collision component of the pragmatic value and the arousal $\mathcal{A}$. At the same time, the decreasing pragmatic value is expressed in an increasingly negative valence $\mathcal{V}$. At $t=\SI{3.4}{s}$ (second dashed line), this causes agent $A$ to change its policy and attempt to avoid the collision by both braking and swerving to the left, as shown in the top panel of Figure~\ref{fig:oncoming_examples}\textbf{b}. However, given the high uncertainty about agent $B$'s behavior, arousal continues to increase, peaking only at $t=\SI{3.8}{s}$ (third dashed line). Only thereafter can agent $A$ become increasingly certain that it has successfully avoided the collision, with $\mathcal{A}$ decreasing and $\mathcal{V}$ increasing. Nevertheless, at $t=\SI{4.6}{s}$ (fourth dashed line), agent $A$ reconsiders its policy, since given the new information, the previous avoidance maneuver was found insufficient for avoiding the collision. The second attempt is more successful, resulting in an immediate drop in arousal.

\subsubsection{Intersection with priority violation.}
\begin{figure*}[!ht]
    \centering
    \includegraphics[scale=0.9]{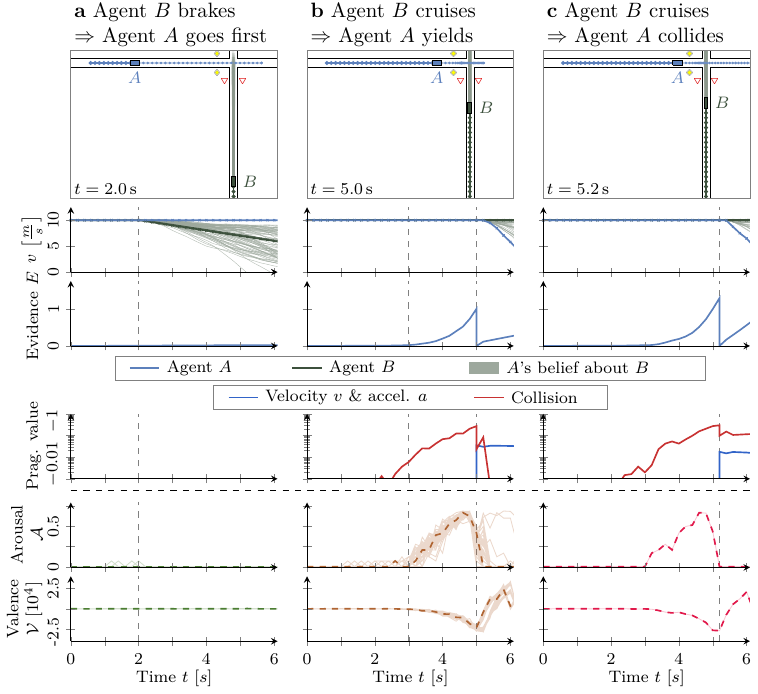}
    \caption{Examples for the three possible outcomes in the intersection scenario. The thin lines in the velocity plot correspond to the prediction shown in the top plots. \textbf{a}) The compliant agent $B$ complies with the priority rules and brakes, allowing agent $A$ to pass first uncontested. \textbf{b}) The non-compliant agent $B$ signals yielding deceptively, but agent $A$ starts to mistrust that signal in time for it to brake itself, yielding to agent $B$ but avoiding a collision. \textbf{c}) In contrast to the previous example, agent $A$ misjudges the danger of the scenario, an brakes to late, resulting in a collision. In the emotion plots (lowest two panels), the dashed lines correspond to the examples shown in the panels above, while the thin lines represent the other simulated cases. The dashed vertical lines correspond to key points in time discussed in the text.}
    \label{fig:conflict_examples}
    \vspace{-0.5cm}
\end{figure*}
In the intersection scenario, we again observe clear differences in the recorded emotions depending both on the behavior of agent $B$ and on the outcome of the scenario. In the scenario with a compliant other agent (Figure~\ref{fig:conflict_examples}\textbf{a}), agent $B$ starts to brake at $t=\SI{2}{s}$ (first dashed line), in accordance with the normative expectations of agent $A$. Based on these expectations, agent $A$ can effectively exclude the possibility of a collision. Consequently, both the arousal $\mathcal{A}$ and the collision contribution to the pragmatic value remain negligible. Furthermore, because this allows agent $A$ to cross the intersection exactly as expected, the valence also remains neutral.

Meanwhile, if agent $B$ ignores the traffic rules and continues towards the intersection at a constant velocity (Figure~\ref{fig:conflict_examples}\textbf{b}), agent $A$'s emotions differ markedly. Specifically, after $t=\SI{3}{s}$ (first dashed line), agent $A$ starts to lose trust in agent $B$'s norm compliance (as this would require increasingly harsh braking), which is reflected in the increasing likelihood of collision outcomes (visible in the higher arousal $\mathcal{A}$ and collision contribution to the pragmatic value). This also results in an increasingly negative valence, as agent $A$'s expectation that agent $B$ will yield is repeatedly violated. Finally, at $t=\SI{5}{s}$ (second dashed line), agent $A$ abandons its priority and begins to brake itself. As a result, the possibility of a collision can largely be disregarded, leading to a drop in $\mathcal{A}$ and an increase in $\mathcal{V}$, since the previously expected collision is no longer anticipated to occur. 

The events with a collision outcome exhibits largely similar emotions compared to the events with a successful resolution (Figure~\ref{fig:conflict_examples}\textbf{c}), with two important differences. First, the decrease in arousal leading up to the braking response by agent $A$ at $t=\SI{5.2}{s}$ (first dashed line) is not caused by the certainty that a collision will be avoided, but instead by the certainty that a collision has become unavoidable. This is reflected in the persistent collision contribution to the pragmatic value thereafter. Second, the valence starts to increase later and reaches a lower peak than in the successful avoidance case. While the decreasing arousal and increasing valence seem counterintuitive, the latter is reasonable given the model's preference for the actual low-speed collision over the initially expected high-speed impact.

\subsubsection{Emotions in the circumplex model.}
\begin{figure}[!ht]
    \centering
    \begin{subfigure}[t]{6cm}
        \centering
        \includegraphics[scale=0.9]{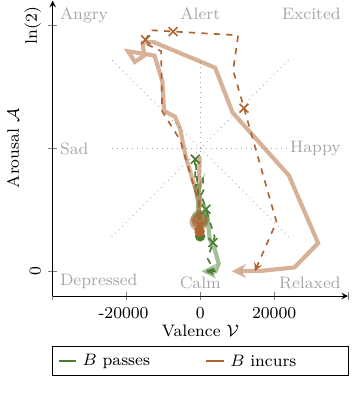}
        \caption{Lateral incursion (Figure~\ref{fig:oncoming_examples}).}
        \label{fig:oncoming_test}
    \end{subfigure}%
    \hfill
    \begin{subfigure}[t]{6cm}
        \centering
        \includegraphics[scale=0.9]{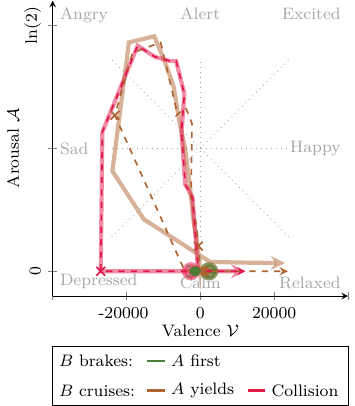}
        \caption{Intersection (Figure~\ref{fig:conflict_examples}).}
        \label{fig:conflict_test}
    \end{subfigure}%
    \caption{The circumplex representation of the emotions for agent $A$ labeled based on Posner et al.~\cite{posner2005circumplex}. The dashed lines correspond to the individual examples shown in Figures~\ref{fig:oncoming_examples} and~\ref{fig:conflict_examples} (with the cross markers showing corresponding to the vertical dashes lines in those figures), while the solid lines show the averaged emotions for each of the outcomes.}
    \label{fig:emotions}
    \vspace{-0.5cm}
\end{figure}
We now map valence $V$ and arousal $\mathcal{A}_B$ onto the circumplex model to infer emotional states directly (Figure~\ref{fig:emotions}). In the lateral incursion scenario, emotions remain relatively stable in the case where the other agent is compliant but exhibit large fluctuations in the non-compliant case (Figure~\ref{fig:oncoming_test}). In particular, agent $A$ initially responds to agent $B$’s norm-violating incursion with anger, before returning to a relaxed state after successfully avoiding the collision.

Emotions are even more stable in the benign intersection scenario, with agent $A$ remaining calm throughout (Figure~\ref{fig:conflict_test}). In the adversarial case, agent $A$ again reacts with anger, which shifts to sadness following the braking maneuver before returning to a relaxed state, reflecting relief after avoiding the collision. By contrast, in the collision scenario, the emotional state transitions from sadness to depression instead.

\section{Discussion}
This paper extends the previous work on modeling emotions with active inference~\cite{pattisapu2024free} to dynamic, interactive driving scenario.
Applied to two traffic scenarios, the resulting model yields affective responses that are consistent with intuitive expectations.
In particular, non-compliant, norm-violating -- and therefore unexpected -- behavior by the other agent induces anger (i.e., negative valence and increased arousal), which is consistent with previous empirical studies~\cite{underwood1999anger,mesken2007frequency}.  
The primary exception is the seemingly counterintuitive decrease in arousal immediately prior to the collision in the intersection scenario. As the model accounts only for the resolved uncertainty over the high-level outcome (i.e., whether a collision occurs), it neglects the persistent uncertainty about the consequences of that collision (e.g., injury or death) that might cause elevated arousal in reality at this stage. Capturing such effects, however, would require explicit modeling of collision mechanics and outcome severity, which is beyond the scope of the present framework.

The observed dependence of emotional responses on traffic dynamics is consistent with prior work~\cite{liu2021empathetic} demonstrating empirically that the behavior of interacting agents is a strong predictor of human affective responses in traffic scenarios. Importantly, our method provides a structured and interpretable account of how these emotional states arise. For instance, by explicitly linking valence to deviations from expected or norm-compliant outcomes, the model offers a plausible mechanistic explanation for the emergence of affect. In this sense, the proposed approach is not only capable of predicting emotional states but also of explaining them in terms of underlying cognitive processes.

Furthermore, our implementation of emotions is consistent regarding appraisal theories of emotion, which emphasize that affective responses are driven by anticipated outcomes of events~\cite{scherer2001appraisal}. This is both the case for arousal being based on uncertainty about the outcome as well as valence being influenced by predicted long-term collision risk rather than immediate observations alone. Moreover, appraisal theories highlight the role of social and normative expectations in shaping emotional responses~\cite{scherer2001appraisal,bediou2014egocentric}. Accordingly, the observed decrease in valence and increase in arousal in response to unexpected or norm-violating behavior is consistent with these mechanisms, as deviations from expected driving norms directly affect evaluative appraisals.

Importantly, our proposed method is in principle applicable to a general class of active inference models with discrete time and continuous state representations where agents optimize policies over multiple predicted time steps. Our approach allows affective quantities to reflect not only instantaneous outcomes of current actions but also to anticipate the risk associated with future trajectories. Furthermore, incorporating information from multiple past policies when estimating valence reduces variability due to stochastic policy selection and state predictions, resulting in more stable affective signals. Together, these extensions demonstrate that emotion modeling can be integrated into a wider range of active inference models without requiring substantial modifications to the underlying framework. Affective states being inferrable from within the same generative framework used to model behavior further support active inference as a unified account of perception, action, and affect~\cite{parr_active_2022}.

However, our modeling effort has the following limitations. First, while our model captures valence as a function of short-term prediction error, this neglects the influence of higher-level or long-term expectations. Intuitively, valence depends not only on immediate deviations from the expected value, but also on sustained discrepancies relative to longer-term expectations. As a result, even in temporally stable environments where short-term prediction errors are minimized, agents may continue to experience negative valence if the current state remains incongruent with prior, long-term expectations~\cite{scherer2001appraisal}. For example, a driver in a traffic jam could still experience negative emotions if they expected the journey to be smooth initially, even if the current scenario is very stable (i.e., marginal prediction errors). Consequently, future work should consider using multiple time-scale, hierarchical models~\cite{pezzulo2018hierarchical} to properly represent these processes.

Second, this work is limited to simulation and thus we have not compared the model predictions to human data. In the future, this could be achieved using existing datasets in which emotions are inferred from facial expressions or manual annotations~\cite{liu2021empathetic,li2021spontaneous,li2022multimodal}. However, such an evaluation would require accurately modeling more complex and realistic interaction dynamics.

Third, the model focuses on the effect of behavior on emotion, while potential effects of emotion on behavior are currently out of its scope. For example, such effects could include emotions modulating the preference prior that shapes pragmatic value. Empirical evidence suggests that negative emotions, such as anger, can lead to increased risk-taking and more self-centered behavior~\cite{li2019angry,yu2022trait}. Within our framework, this could be captured by reducing penalties for collisions or norm violations under negative valence. Similar bidirectional interactions between internal states and behavior have been explored in prior work (e.g., Schwarting et al.~\cite{schwarting_social_2019}), and could be a natural extension of the present approach.

To conclude, we proposed a novel method for quantifying agents' emotional states in an active inference model of dynamic and interactive human driving. Following the previously proposed circumplex model of human emotion, we infer valence and arousal of an active inference agent based on kinematic observations alone. In simulations, we showed the promise of this approach, demonstrating that our agent tends to react negatively to unexpected norm violations by other agents, qualitatively consistent with previous studies involving human drivers. Our work offers a significant stepping stone toward better understanding human interactions in traffic, while enabling a more seamless and acceptable integration of autonomous vehicles into mixed traffic.

\bibliographystyle{splncs04}
\bibliography{manual_bib}
\end{document}